\documentclass[11pt,a4paper]{article}
\usepackage[hyperref]{emnlp-ijcnlp-2019}
\usepackage{times}
\usepackage{latexsym}

\usepackage{amsmath}
\usepackage{amssymb}
\usepackage{amsfonts}
\usepackage{graphicx}
\usepackage{soul}
\usepackage{multirow}
\usepackage{url}
\usepackage{makecell}
\usepackage[normalem]{ulem}
\usepackage{url}
\usepackage{comment}
\usepackage{adjustbox}
\newcolumntype{L}[1]{>{\raggedright\let\newline\\\arraybackslash\hspace{0pt}}m{#1}}

\aclfinalcopy 

\newcommand*{\affaddr}[1]{#1} 
\newcommand*{\affmark}[1][*]{\textsuperscript{#1}}
\newcommand*{\email}[1]{\texttt{#1}}

\title{Exploiting BERT for End-to-End Aspect-based Sentiment Analysis\thanks{\hspace{0.15cm}The work described in this paper is substantially supported by a grant from the Research Grant Council of the Hong Kong Special Administrative Region, China (Project Code: 14204418).}}

\author{%
Xin Li\affmark[1], Lidong Bing\affmark[2], Wenxuan Zhang\affmark[1] and Wai Lam\affmark[1] \\
\affaddr{\affmark[1]Department of Systems Engineering and Engineering Management\\
The Chinese University of Hong Kong, Hong Kong}\\
\affaddr{\affmark[2]R\&D Center Singapore, Machine Intelligence Technology, Alibaba DAMO Academy}\\
\email{\{lixin,wxzhang,wlam\}@se.cuhk.edu.hk}\\
\email{l.bing@alibaba-inc.com}\\
}

\date{}

\begin{document}
\maketitle
\begin{abstract}
In this paper, we investigate the modeling power of contextualized embeddings from pre-trained language models, e.g. BERT, on the E2E-ABSA task. Specifically, we build a series of simple yet insightful neural baselines to deal with E2E-ABSA. The experimental results show that even with a simple linear classification layer, our BERT-based architecture can outperform state-of-the-art works. Besides, we also standardize the comparative study by consistently utilizing a hold-out development dataset for model selection, which is largely ignored by previous works. Therefore, our work can serve as a BERT-based benchmark for E2E-ABSA.\footnote{Our code is open-source and available at: \url{https://github.com/lixin4ever/BERT-E2E-ABSA}}
\end{abstract}

\section{Introduction}
Aspect-based sentiment analysis (ABSA) is to discover the users' sentiment or opinion towards an aspect, usually in the form of explicitly mentioned aspect terms~\cite{mitchell-etal-2013-open,zhang-etal-2015-neural} or implicit aspect categories~\cite{wang-etal-2016-attention}, from user-generated natural language texts~\cite{liu2012sentiment}. The most popular ABSA benchmark datasets are from SemEval ABSA challenges~\cite{pontiki-etal-2014-semeval,pontiki-etal-2015-semeval,pontiki-etal-2016-semeval} where a few thousand review sentences with gold standard aspect sentiment annotations are provided. 

Table~\ref{tab:problem_settings} summarizes three existing research problems related to ABSA. The first one is the original ABSA, aiming at predicting the sentiment polarity of the sentence towards the given aspect. Compared to this classification problem, the second one and the third one, namely, Aspect-oriented Opinion Words Extraction (AOWE)~\cite{fan-etal-2019-target} and End-to-End Aspect-based Sentiment Analysis (E2E-ABSA)~\cite{ma-etal-2018-joint,schmitt-etal-2018-joint,li2019unified,li2017learning,li2019learning}, are related to a sequence tagging problem. Precisely, the goal of AOWE is to extract the aspect-specific opinion words from the sentence given the aspect. The goal of E2E-ABSA is to jointly detect aspect terms/categories and the corresponding aspect sentiments.  

Many neural models composed of a task-agnostic pre-trained word embedding layer and task-specific neural architecture have been proposed for the original ABSA task (i.e. the aspect-level sentiment classification)~\cite{tang-etal-2016-aspect,wang-etal-2016-attention,chen-etal-2017-recurrent-attention,liu-zhang-2017-attention,ma2017interactive,ma2018targeted,majumder-etal-2018-iarm,li-etal-2018-transformation,he-etal-2018-exploiting,xue-li-2018-aspect,wang-etal-2018-target,fan-etal-2018-multi,huang-carley-2018-parameterized,lei2019human,li2019exploiting,zhang2019aspect}\footnote{Due to the limited space, we can not list all of the existing works here, please refer to the survey~\cite{zhou2019deep} for more related papers.}, but the improvement of these models measured by the accuracy or F1 score has reached a bottleneck. One reason is that the task-agnostic embedding layer, usually a linear layer initialized with Word2Vec~\cite{mikolov2013distributed} or GloVe~\cite{pennington-etal-2014-glove}, only provides context-independent word-level features, which is insufficient for capturing the complex semantic dependencies in the sentence. Meanwhile, the size of existing datasets is too small to train sophisticated task-specific architectures. Thus, introducing a context-aware word embedding\footnote{In this paper, we generalize the concept of ``word embedding'' as a mapping between the word and the low-dimensional word representations.} layer pre-trained on large-scale datasets with deep LSTM~\cite{mccann2017learned,peters-etal-2018-deep,howard-ruder-2018-universal} or Transformer~\cite{radford2018improving,radford2019language,devlin-etal-2019-bert,lample2019cross,yang2019xlnet,dong2019unified} for fine-tuning a lightweight task-specific network using the labeled data has good potential for further enhancing the performance. 

\begin{table}[]
    \centering
    \resizebox{1.0\columnwidth}{!}{
    \begin{tabular}{lll}
        \multirow{2}{*}{Sentence:} & \multicolumn{2}{l}{ \texttt{\textbf{<}\ul{Great}\textbf{>} \textbf{[}\textbf{\ul{food}}\textbf{]}$_{\mathrm{P}}$ but the}}  \\
        & \multicolumn{2}{l}{\textbf{[}\textbf{\uwave{service}}\textbf{]}$_{\mathrm{N}}$  \texttt{is \textbf{<}\uwave{dreadful}\textbf{>}.}} \\
        \Xhline{3\arrayrulewidth}
        Settings & Input & Output \\ \hline \hline
        1. ABSA & sentence, aspect & aspect sentiment \\
        2. AOWE & sentence, aspect & opinion words \\
        3. E2E-ABSA & sentence & aspect, aspect sentiment \\
        \Xhline{3\arrayrulewidth}
    \end{tabular}}
    \caption{Different problem settings in ABSA. Gold standard aspects and opinions are wrapped in \texttt{\textbf{[]}} and \texttt{\textbf{<>}} respectively. The subscripts N and P refer to aspect sentiment. Underline \uwave{*} or \ul{*} indicates the association between the aspect and the opinion.}
    \label{tab:problem_settings}
\end{table}

\citet{xu-etal-2019-bert,sun-etal-2019-utilizing,song2019attentional,yu2019adapting,rietzler2019adapt,huang2019syntax,hu2019learning} have conducted some initial attempts to couple the deep contextualized word embedding layer with downstream neural models for the original ABSA task and establish the new state-of-the-art results. It encourages us to explore the potential of using such contextualized embeddings to the more difficult but practical task, i.e. E2E-ABSA (the third setting in Table~\ref{tab:problem_settings}).\footnote{Both of ABSA and AOWE assume that the aspects in a sentence are given. Such setting makes them less practical in real-world scenarios since manual annotation of the fine-grained aspect mentions/categories is quite expensive.}
Note that we are not aiming at developing a task-specific architecture, instead, our focus is to examine the potential of contextualized embedding for E2E-ABSA, coupled with various simple layers for prediction of E2E-ABSA labels.\footnote{\citet{hu-etal-2019-open} introduce BERT to handle the E2E-ABSA problem but their focus is to design a task-specific architecture rather than exploring the potential of BERT.}


In this paper, we investigate the modeling power of BERT~\cite{devlin-etal-2019-bert}, one of the most popular pre-trained language model armed with Transformer~\cite{vaswani2017attention}, on the task of E2E-ABSA. Concretely, inspired by the investigation of E2E-ABSA in~\citet{li2019unified}, which predicts aspect boundaries as well as aspect sentiments using a single sequence tagger, we build a series of simple yet insightful neural baselines for the sequence labeling problem and fine-tune the task-specific components with BERT or deem BERT as feature extractor. Besides, we standardize the comparative study by consistently utilizing the hold-out development dataset for model selection, which is ignored in most of the existing ABSA works~\cite{tay2018learning}. 

\section{Model}
\begin{figure}
    \centering
    \includegraphics[width=0.5\textwidth]{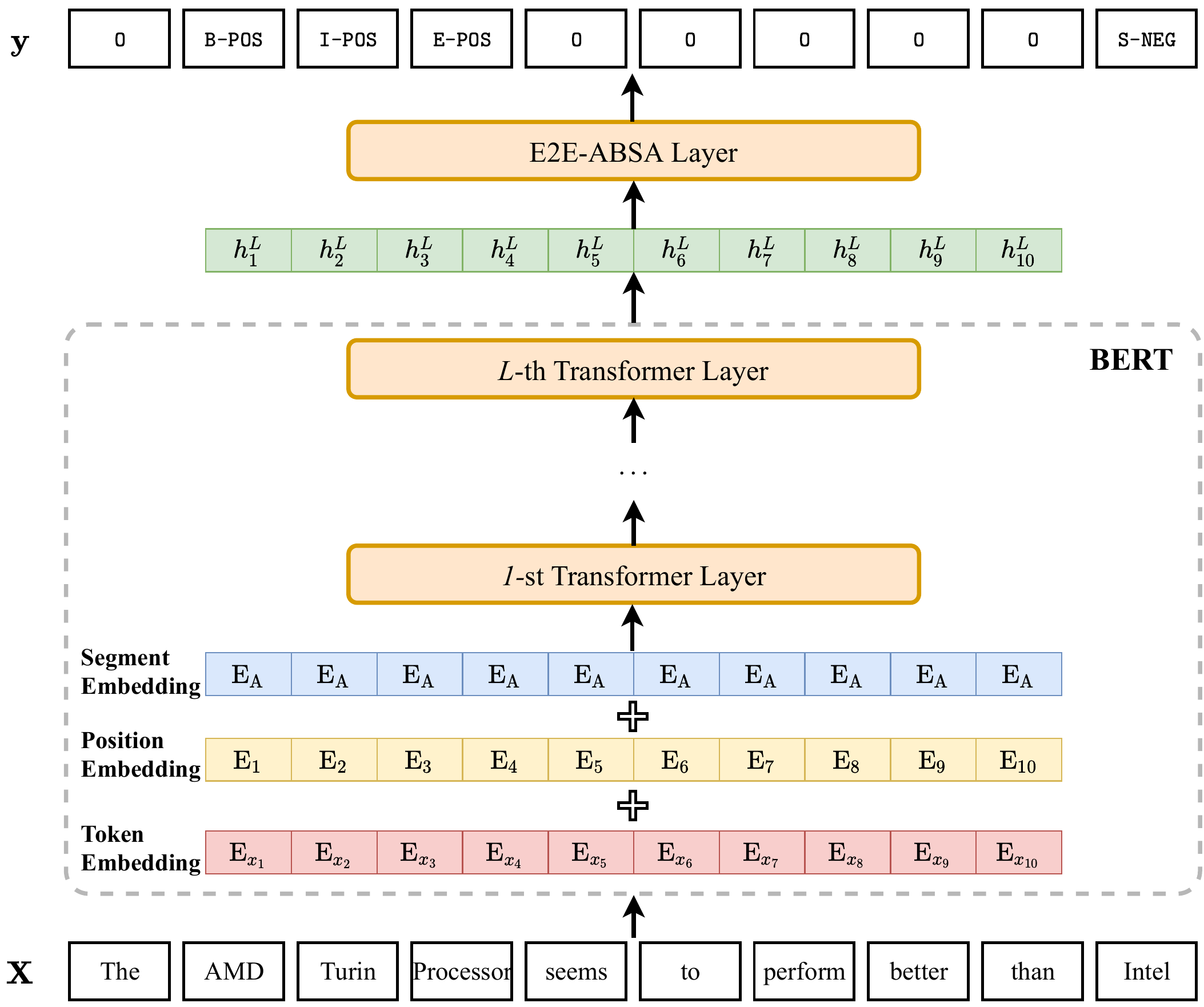}
    \caption{Overview of the designed model.}
    \label{fig:architecture}
\end{figure}

In this paper, we focus on the aspect term-level End-to-End Aspect-Based Sentiment Analysis (E2E-ABSA) problem setting. This task can be formulated as a sequence labeling problem. The overall architecture of our model is depicted in Figure~\ref{fig:architecture}. Given the input token sequence $\mathrm{\bf x} = \{x_1, \cdots, x_T \}$ of length $T$, we firstly employ BERT component with $L$ transformer layers to calculate the corresponding contextualized representations $H^{L} = \{h^L_1, \cdots, h^L_T\} \in \mathbb{R}^{T \times \mathrm{dim}_h}$ for the input tokens where $\mathrm{dim}_h$ denotes the dimension of the representation vector. Then, the contextualized representations are fed to the task-specific layers to predict the tag sequence $\mathrm{\bf y} = \{y_1, \cdots, y_T\}$. The possible values of the tag $y_t$ are \texttt{B}-\{\texttt{POS,NEG,NEU}\}, \texttt{I}-\{\texttt{POS,NEG,NEU}\}, \texttt{E}-\{\texttt{POS,NEG,NEU}\}, \texttt{S}-\{\texttt{POS,NEG,NEU}\} or \texttt{O}, denoting the beginning of aspect, inside of aspect, end of aspect, single-word aspect, with positive, negative or neutral sentiment respectively, as well as outside of aspect.

\subsection{BERT as Embedding Layer}
Compared to the traditional Word2Vec- or GloVe-based embedding layer which only provides a single context-independent representation for each token, the BERT embedding layer takes the sentence as input and calculates the token-level representations using the information from the entire sentence. First of all, we pack the input features as $H^0 = \{e_1,\cdots,e_T\}$, where $e_t$ ($t\in[1, T]$) is the combination of the token embedding, position embedding and segment embedding corresponding to the input token $x_t$. Then $L$ transformer layers are introduced to refine the token-level features layer by layer. Specifically, the representations $H^{l} = \{h^l_1, \cdots, h^l_T\}$ at the $l$-th ($l \in [1, L]$) layer are calculated below:
\begin{equation}
\label{eq:bert}
    H^l = \text{Transformer}_l(H^{l-1})
\end{equation}
We regard $H^L$ as the contextualized representations of the input tokens and use them to perform the predictions for the downstream task. 


\subsection{Design of Downstream Model}
After obtaining the BERT representations, we design a neural layer, called E2E-ABSA layer in Figure~\ref{tab:problem_settings}, on top of BERT embedding layer for solving the task of E2E-ABSA. We investigate several different design for the E2E-ABSA layer, namely, linear layer, recurrent neural networks, self-attention networks, and conditional random fields layer. 
\paragraph{Linear Layer} The obtained token representations can be directly fed to linear layer with softmax activation function to calculate the token-level predictions:
\begin{equation}
    P(y_t|x_t) = \text{softmax}(W_o h^L_t + b_o) 
\end{equation}
where $W_o \in \mathbb{R}^{\mathrm{dim}_h \times |\mathcal{Y}|}$ is the learnable parameters of the linear layer. 
\paragraph{Recurrent Neural Networks} Considering its sequence labeling formulation, Recurrent Neural Networks (RNN)~\cite{elman1990finding} is a natural solution for the task of E2E-ABSA. In this paper, we adopt GRU~\cite{cho-etal-2014-learning}, whose superiority compared to LSTM~\cite{hochreiter1997long} and basic RNN has been verified in~\citet{jozefowicz2015empirical}. The computational formula of the task-specific hidden representation $h^{\mathcal{T}}_t \in \mathbb{R}^{\mathrm{dim}_h}$ at the $t$-th time step is shown below:
\begin{equation}
\small
    \begin{split}
        \begin{bmatrix}
        r_t \\
        z_t
     \end{bmatrix}
     &= \sigma(\textsc{Ln}(W_x h^L_t)+\textsc{Ln}(W_h h^{\mathcal{T}}_{t-1})) \\
     n_t &= \text{tanh}(\textsc{Ln}(W_{xn} h^L_t)+r_t * \textsc{Ln}(W_{hn} h^{\mathcal{T}}_{t-1})) \\
     h^{\mathcal{T}}_{t} &= (1-z_t) * n_t + z_t * h^{\mathcal{T}}_{t-1}
    \end{split}
\end{equation}
where $\sigma$ is the sigmoid activation function and $r_t$, $z_t$, $n_t$ respectively denote the reset gate, update gate and new gate. $W_x,W_h \in \mathbb{R}^{2\mathrm{dim}_h \times \mathrm{dim}_h}$, $W_{xn}, W_{hn} \in \mathbb{R}^{\mathrm{dim}_h \times \mathrm{dim}_h}$ are the parameters of GRU. Since directly applying RNN on the output of transformer, namely, the BERT representation $h^L_t$, may lead to unstable training~\cite{chen-etal-2018-best,liu2019fine}, we add additional layer-normalization~\cite{ba2016layer}, denoted as $\textsc{Ln}$, when calculating the gates. Then, the predictions are obtained by introducing a softmax layer:
\begin{equation}
\label{output}
    p(y_t|x_t) = \text{softmax}(W_o h^{\mathcal{T}}_t + b_o) 
\end{equation}

\paragraph{Self-Attention Networks} With the help of self attention~\cite{cheng-etal-2016-long,lin2017structured}, Self-Attention Network~\cite{vaswani2017attention,shen2018disan} is another effective feature extractor apart from RNN and CNN. In this paper, we introduce two SAN variants to build the task-specific token representations $H^{\mathcal{T}}=\{h^{\mathcal{T}}_1,\cdots,h^{\mathcal{T}}_T\}$. One variant is composed of a simple self-attention layer and residual connection~\cite{he2016deep}, dubbed as ``SAN''. The computational process of SAN is below:
\begin{equation}
\begin{split}
    H^{\mathcal{T}} &= \textsc{Ln}(H^{L}+\textsc{Slf-Att} (Q, K, V)) \\
    Q, K, V &=  H^{L} W^{Q}, H^{L}W^{K}, H^{L} W^{V}
\end{split}
\end{equation}
where \textsc{Slf-Att} is identical to the self-attentive scaled dot-product attention~\cite{vaswani2017attention}. Another variant is a transformer layer (dubbed as ``TFM''), which has the same architecture with the transformer encoder layer in the BERT. The computational process of TFM is as follows: 
\begin{equation}
\begin{split}
    \hat{H}^{L} &= \textsc{Ln}(H^{L}+\textsc{Slf-Att} (Q, K, V)) \\
    H^{\mathcal{T}} &= \textsc{Ln}(\hat{H}^{L}+\textsc{Ffn}(\hat{H}^{L}))
\end{split}
\end{equation}
where \textsc{Ffn} refers to the point-wise feed-forward networks~\cite{vaswani2017attention}. Again, a linear layer with softmax activation is stacked on the designed SAN/TFM layer to output the predictions (same with that in Eq(\ref{output})).

\begin{table*}[]
    \centering
    \resizebox{0.80\textwidth}{!}{
    \begin{tabular}{ll|ccc|ccc}
    \Xhline{3\arrayrulewidth}
        & \multirow{2}{*}{Model} & \multicolumn{3}{c|}{\texttt{LAPTOP}} & \multicolumn{3}{c}{\texttt{REST}}  \\ \cline{3-8}
        & & P & R & F1 & P & R & F1 \\ \hline \hline
        \multirow{3}{*}{Existing Models} 
       & \cite{li2019unified}$^{\natural}$ & 61.27 & 54.89 & 57.90 & 68.64 & 71.01 & 69.80 \\
       & \cite{luo-etal-2019-doer}$^{\natural}$ & - & - & 60.35 & - & - & 72.78 \\
       & \cite{he-etal-2019-interactive}$^{\natural}$ & - & - & 58.37 & - & - & - \\  \hline
       \multirow{3}{*}{LSTM-CRF} & \cite{lample-etal-2016-neural}$^{\sharp}$ & 58.61 & 50.47 & 54.24 & 66.10 & 66.30 & 66.20\\
       & \cite{ma-hovy-2016-end}$^{\sharp}$ & 58.66 & 51.26 & 54.71 & 61.56 & 67.26 & 64.29 \\
       & \cite{liu2018empower}$^{\sharp}$ & 53.31 & 59.40 & 56.19 & 68.46 & 64.43 & 66.38 \\ \hline
       \multirow{5}{*}{BERT Models} & BERT+Linear & 62.16 & 58.90 & 60.43 & 71.42 & 75.25 & 73.22 \\
       & BERT+GRU & 61.88 & 60.47 & \textbf{61.12} & 70.61 & 76.20 & 73.24 \\
       & BERT+SAN & 62.42 & 58.71 & 60.49 & 72.92 & 76.72 & \textbf{74.72} \\
       & BERT+TFM & 63.23 & 58.64 & 60.80 & 72.39 & 76.64 & 74.41 \\
       & BERT+CRF & 62.22 & 59.49 & 60.78 & 71.88 & 76.48 & 74.06 \\
    \Xhline{3\arrayrulewidth}
    \end{tabular}}
    \caption{Main results. The symbol $^{\natural}$ denotes the numbers are officially reported ones. The results with $^{\sharp}$ are retrieved from~\citet{li2019unified}.}
    \label{tab:main_results}
\end{table*}

\begin{table}[]
    \centering
    \begin{small}
    \begin{tabular}{ll|c|c|c|c}
    \Xhline{2\arrayrulewidth}
        \multicolumn{2}{c|}{Dataset} & Train & Dev & Test & Total  \\ \hline
         \multirow{2}{*}{\texttt{LAPTOP}} & \# sent & 2741 & 304 & 800 & 4245 \\ 
        & \# aspect & 2041 & 256 & 634 & 2931 \\ \hline
        \multirow{2}{*}{\texttt{REST}} & \# sent & 3490 & 387 & 2158 & 6035 \\ 
        & \# aspect & 3893 & 413 & 2287 & 6593 \\ 
    \Xhline{2\arrayrulewidth}
    \end{tabular}
    \end{small}
    \caption{Statistics of datasets.}
    \label{tab:dataset}
\end{table}

\paragraph{Conditional Random Fields} Conditional Random Fields (CRF)~\cite{lafferty2001conditional} is effective in sequence modeling and has been widely adopted for solving the sequence labeling tasks together with neural models~\cite{huang2015bidirectional,lample-etal-2016-neural,ma-hovy-2016-end}. In this paper, we introduce a linear-chain CRF layer on top of the BERT embedding layer. Different from the above mentioned neural models maximizing the token-level likelihood $p(y_t|x_t)$, the CRF-based model aims to find the globally most probable tag sequence. Specifically, the sequence-level scores $s(\mathrm{\bf x},\mathrm{\bf y})$ and likelihood $p(\mathrm{\bf y}|\mathrm{\bf x})$ of $\mathrm{\bf y} = \{y_1,\cdots,y_T\}$ are calculated as follows:
\begin{equation}
    \begin{split}
        s(\mathrm{\bf x},\mathrm{\bf y}) &= \sum^T_{t=0} M^A_{y_t,y_{t+1}} + \sum^T_{t=1} M^P_{t, y_t} \\
        p(\mathrm{\bf y}|\mathrm{\bf x}) &= \text{softmax}(s(\mathrm{\bf x},\mathrm{\bf y}))
    \end{split}
\end{equation}
where $M^A \in \mathbb{R}^{|\mathcal{Y}| \times |\mathcal{Y}|}$ is the randomly initialized transition matrix for modeling the dependency between the adjacent predictions and $M^P \in \mathbb{R}^{T \times |\mathcal{Y}|}$ denote the emission matrix linearly transformed from the BERT representations $H^L$. The softmax here is conducted over all of the possible tag sequences. As for the decoding, we regard the tag sequence with the highest scores as output:
\begin{equation}
    \mathrm{\bf y}^* = \arg\max_{\mathrm{\bf y}} s(\mathrm{\bf x},\mathrm{\bf y})
\end{equation}
where the solution is obtained via Viterbi search.

\begin{figure}
\centering
    \includegraphics[width=0.40\textwidth]{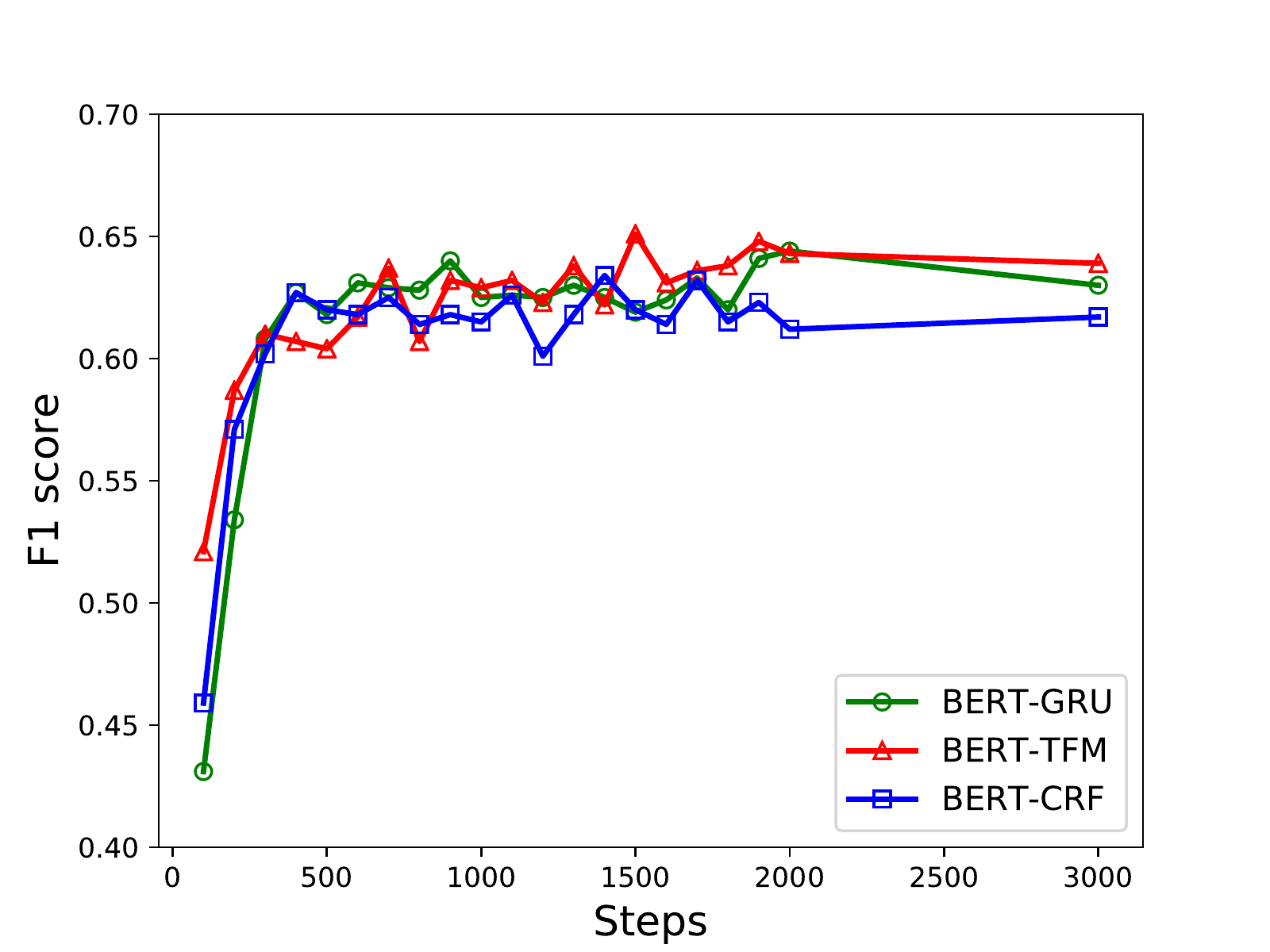}
    \caption{Performances on the Dev set of \texttt{REST}.}
    \label{fig:overfit}
\end{figure}

\section{Experiment}
\subsection{Dataset and Settings}
We conduct experiments on two review datasets originating from SemEval~\cite{pontiki-etal-2014-semeval,pontiki-etal-2015-semeval,pontiki-etal-2016-semeval} but re-prepared in~\citet{li2019unified}. The statistics are summarized in Table~\ref{tab:dataset}. 
We use the pre-trained ``bert-base-uncased'' model\footnote{https://github.com/huggingface/transformers}, where the number of transformer layers $L=12$ and the hidden size $\mathrm{dim}_h$ is 768. For the downstream E2E-ABSA component, we consistently use the single-layer architecture and set the dimension of task-specific representation as $\mathrm{dim}_h$. The learning rate is 2e-5. The batch size is set as 25 for \texttt{LAPTOP} and 16 for \texttt{REST}. We train the model up to 1500 steps. After training 1000 steps, we conduct model selection on the development set for very 100 steps according to the micro-averaged F1 score. Following these settings, we train 5 models with different random seeds and report the average results.


We compare with \textbf{Existing Models}, including tailor-made E2E-ABSA models~\cite{li2019unified,luo-etal-2019-doer,he-etal-2019-interactive}, and competitive \textbf{LSTM-CRF} sequence labeling models~\cite{lample-etal-2016-neural,ma-hovy-2016-end,liu2018empower}.

\subsection{Main Results}
From Table~\ref{tab:main_results}, we surprisingly find that only introducing a simple token-level classifier, namely, BERT-Linear, already outperforms the existing works without using BERT, suggesting that BERT representations encoding the associations between arbitrary two tokens largely alleviate the issue of context independence in the linear E2E-ABSA layer. It is also observed that slightly more powerful E2E-ABSA layers lead to much better performance, verifying the postulation that incorporating context helps to sequence modeling.

\subsection{Over-parameterization Issue}
Although we employ the smallest pre-trained BERT model, it is still over-parameterized for the E2E-ABSA task (110M parameters), which naturally raises a question: does BERT-based model tend to overfit the small training set? Following this question, we train BERT-GRU, BERT-TFM and BERT-CRF up to 3000 steps on \texttt{REST} and observe the fluctuation of the F1 measures on the development set. As shown in Figure~\ref{fig:overfit}, F1 scores on the development set are quite stable and do not decrease much as the training proceeds, which shows that the BERT-based model is exceptionally robust to overfitting. 

\subsection{Finetuning BERT or Not}
\begin{figure}
    \centering
    \includegraphics[width=0.4\textwidth]{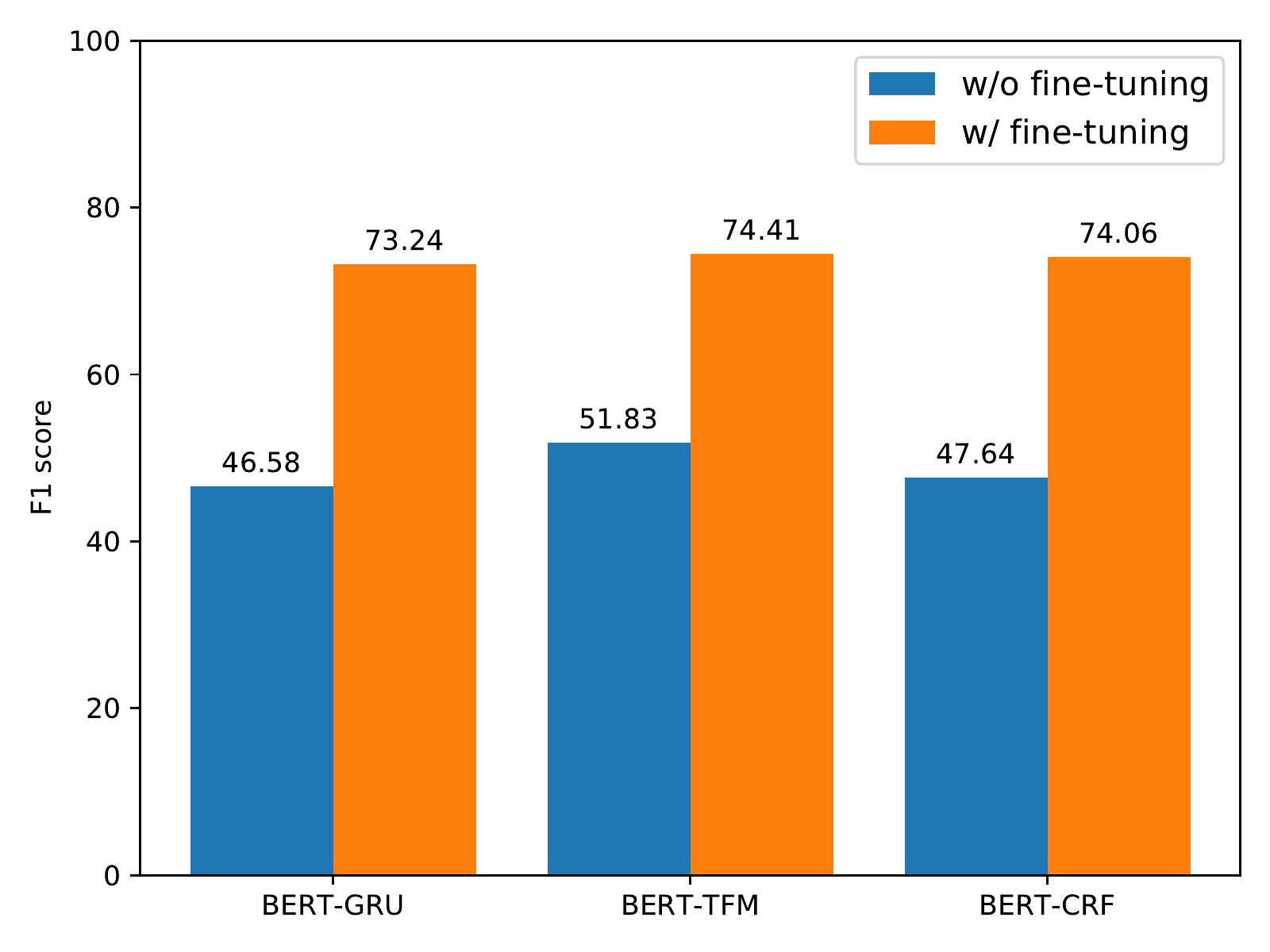}
    \caption{Effect of fine-tuning BERT.}
    \label{fig:finetune}
\end{figure}

We also study the impact of fine-tuning on the final performances. Specifically, we employ BERT to calculate the contextualized token-level representations but kept the parameters of BERT component unchanged in the training phase. Figure~\ref{fig:finetune} illustrate the comparative results between the BERT-based models and those keeping BERT component fixed. Obviously, the general purpose BERT representation is far from satisfactory for the downstream tasks and task-specific fine-tuning is essential for exploiting the strengths of BERT to improve the performance.

\section{Conclusion}
In this paper, we investigate the effectiveness of BERT embedding component on the task of End-to-End Aspect-Based Sentiment Analysis (E2E-ABSA). Specifically, we explore to couple the BERT embedding component with various neural models and conduct extensive experiments on two benchmark datasets. The experimental results demonstrate the superiority of BERT-based models on capturing aspect-based sentiment and their robustness to overfitting.

\bibliographystyle{acl_natbib}
\bibliography{emnlp-ijcnlp-2019}

\end{document}